\documentclass[conference]{IEEEtran}
\IEEEoverridecommandlockouts
\usepackage{cite}
\usepackage{url}
\usepackage{kotex}
\usepackage{amsmath,amssymb,amsfonts}
\usepackage{algorithm, algpseudocode}

\usepackage{graphicx}
\usepackage{textcomp}
\usepackage{xcolor}
\usepackage{subfigure}

\usepackage{hyperref}
\hypersetup{
    colorlinks=true,
    linkcolor=blue,
    filecolor=magenta,      
    urlcolor=cyan,
}

\def\BibTeX{{\rm B\kern-.05em{\sc i\kern-.025em b}\kern-.08em
    T\kern-.1667em\lower.7ex\hbox{E}\kern-.125emX}}
\begin{document}

\title{InfoFusion Controller: Informed TRRT Star with Mutual Information based on Fusion of Pure Pursuit and MPC for Enhanced Path Planning \\
\thanks{*This research was supported by Brian Impact Foundation, a non-profit organization dedicated to the advancement of science and technology for all}
}

\author{
\IEEEauthorblockN{Seongjun Choi}
\IEEEauthorblockA{\textit{Kyung-Hee University} \\
Autonomous Driving Lab, \\
MODULABS, Republic of Korea \\}
\and
\IEEEauthorblockN{Youngbum Kim}
\IEEEauthorblockA{\textit{Korea Aviation University} \\
Autonomous Driving Lab, \\
MODULABS, Republic of Korea \\}
\and
\IEEEauthorblockN{Nam Woo Kim}
\IEEEauthorblockA{\textit{Unity Technologies} \\
Autonomous Driving Lab, \\
MODULABS, Republic of Korea \\}
\and
\IEEEauthorblockN{Mansun Shin}
\IEEEauthorblockA{\textit{SPACEEDUING Co., Ltd.} \\
Autonomous Driving Lab, \\
MODULABS, Republic of Korea \\}
\and
\IEEEauthorblockN{Byunggi Chae}
\IEEEauthorblockA{\textit{Auroka Pankyo} \\
Autonomous Driving Lab, \\
MODULABS, Republic of Korea \\}
\and
\IEEEauthorblockN{Sungjin Lee}
\IEEEauthorblockA{\textit{Dong Seoul University,} \\
Autonomous Driving Lab, \\
MODULABS, Republic of Korea \\}
\thanks{\textsuperscript{*}Seongjun Choi and Youngbum Kim are co-first auther.}
}

\maketitle
\begin{abstract} 


In this paper, we propose the \textit{InfoFusion Controller}, an advanced path planning algorithm that integrates both global and local planning strategies to enhance autonomous driving in complex urban environments. The global planner utilizes the informed Theta-Rapidly-exploring Random Tree Star (Informed-TRRT*) algorithm to generate an optimal reference path, while the local planner combines Model Predictive Control (MPC) and Pure Pursuit algorithms. Mutual Information (MI) is employed to fuse the outputs of the MPC and Pure Pursuit controllers, effectively balancing their strengths and compensating for their weaknesses.

The proposed method addresses the challenges of navigating in dynamic environments with unpredictable obstacles by reducing uncertainty in local path planning and improving dynamic obstacle avoidance capabilities. Experimental results demonstrate that the InfoFusion Controller outperforms traditional methods in terms of safety, stability, and efficiency across various scenarios, including complex maps generated using SLAM techniques.

The code for the InfoFusion Controller is available at \url{https://github.com/DrawingProcess/InfoFusionController}. 
\end{abstract}


\section{Introduction} 
\label{sec:intro}

Due to recent developments in artificial intelligence and electric vehicle platform technologies, autonomous driving technology has advanced from Level 2 and Level 3 towards Level 4 and Level 5 automation \cite{SAE:J3016}. The autonomous driving technologies have the objective of decreasing traffic accidents resulting from human error and freeing individuals from the task of driving, ultimately offering a safer and more comfortable driving experience \cite{Li:2024}. Despite the considerable advancements in autonomous driving technology, there are ongoing reservations about the effectiveness and safety of autonomous vehicles in urban settings, particularly in high-traffic situations \cite{Matsui:2022, MacDonald:2016, Bai:2021, Yin:2023}.

This paper proposes a method that combines global and local path planning to achieve both global route optimization and real-time obstacle avoidance in complex environments. Specifically, it enhances dynamic obstacle avoidance performance by optimizing local path planning using MI.

The contributions of this paper are summarized as follows:
\begin{itemize}
    \item Algorithms suitable for global and local path planning: An algorithm for global and local path planning that enables simultaneous global optimization and real-time obstacle avoidance in complex environments. 
    \item Optimization of local path planning using MI: Local path optimization using MI to reduce uncertainty and ensure safe, real-time path adjustments. 
    \item Improvement in dynamic obstacle avoidance: Enhanced dynamic obstacle avoidance through precise prediction and real-time path recalibration.
\end{itemize}

\begin{figure*} [!h]
\centering
\includegraphics[width=\linewidth]{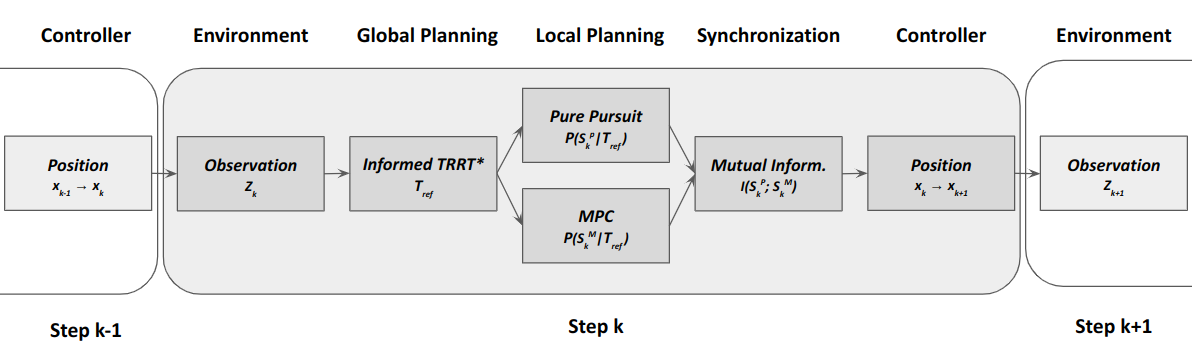}
\caption{Overall Operation of the Proposed Path Planning Algorithm: The diagram illustrates the complete operation of the proposed path planning algorithm, encompassing both the Global Planning and Local Planning stages.}\label{fig:Architecture}
\end{figure*}


\section{Related Works} 
\label{sec:rel}

There are two main components in path planning for autonomous vehicles: Global Path Planning and Local Path Planning, each presenting distinct strengths and weaknesses. Numerous studies have suggested a range of strategies to tackle these obstacles in path planning.

This research \cite{Yang:2022} introduced a method for path planning in autonomous vehicles that is based on risk assessment, enabling the vehicle to select a safer route by analyzing potential risks along the way. Another research \cite{Matsui:2022} proposed the utilization of hierarchical maps for the purposes of global and local path planning in autonomous robots.

Furthermore, this research \cite{Daniel:2010} presented the Informed RRT* algorithm, a sampling-based technique designed for optimizing path planning in global environments. This method demonstrates more efficient sampling capabilities compared to the traditional RRT* algorithm, resulting in a reduction of the time needed for path planning. These researches \cite{Shi:2024, Lee:2024} enhanced the Informed-TRRT* algorithm to optimize path finding in intricate environments, resulting in increased efficiency.

This research \cite{Li:2023} introduced the Adaptive Pure Pursuit algorithm, which facilitates precise path tracking during real-time path planning. This research \cite{Singh:2024} proposed a hybrid approach to path planning aimed at optimizing the navigation efficiency of unmanned vehicles. This technology improves the capability to effectively establish routes in unrestricted settings, expanding its potential usefulness across diverse scenarios.

Significantly, the research \cite{MacDonald:2016} MI to suggest a reactive motion planning strategy based on MI policies in environments with uncertainty. This research describes a technique for self-driving vehicles to analyze real-time environmental data and make corresponding adjustments to their trajectories. Path planning utilizing machine intelligence focuses on integrating multiple sources of sensor data in order to measure uncertainty and establish paths that are more dependable \cite{Hadizadeh:2024, He:2020, Kopylova:2008, Bai:2021}.

%
%
%
%

\section{Methods}
\label{sec:methods}

\subsection{Architecture Overview}

The architecture of Figure \ref{fig:Architecture} illustrates the flow of path planning and control in a robot or autonomous vehicle system. The process is divided into distinct stages, each representing interactions with the environment and the application of various planning algorithms to compute the vehicle's subsequent position.

\begin{itemize}
    \item \textbf{Controller - Step k-1}:
    The vehicle's current position $x_{k-1}$ is given, and this positional information is passed to the environment.

    \item \textbf{Environment - Observation}:
    The environment provides observational data $Z_k$, which includes information about obstacles, road conditions, and other environmental factors affecting the vehicle.

    \item \textbf{Global Planning - Informed TRRT*}:
    In the global planning stage, the \textbf{Informed TRRT*} algorithm is used to generate an optimal reference path $T_{\text{ref}}$ based on the observation $Z_k$ and current state $x_k$. This reference path is then passed to the local planning stage.

    \item \textbf{Local Planning}:
    The local planning stage uses two primary algorithms: \textbf{Pure Pursuit} and \textbf{MPC}.
    \begin{itemize}
        \item \textbf{Pure Pursuit}: This algorithm simplifies the vehicle's movement control by steering it towards a point on the reference path $T_{\text{ref}}$.
        \item \textbf{MPC}: This algorithm predicts the vehicle's optimal control inputs by considering a dynamic model over a prediction horizon, optimizing to follow the reference path.
    \end{itemize}

    \item \textbf{Synchronization - MI}:
    To combine the outputs of the two local control algorithms (MPC and Pure Pursuit), \textbf{MI} is employed. The MI between the two sets of state predictions $S_k^P$ (Pure Pursuit) and $S_k^M$ (MPC) is calculated, helping select the most suitable path.

    \item \textbf{Controller - Step k+1}:
    After selecting the optimal path based on the MI, the vehicle proceeds to the next state $x_{k+1}$.
\end{itemize}
This architecture provides a structured explanation of how an autonomous driving system interacts with the environment in real-time to select the optimal path and reliably control the vehicle.

\subsection{MPC}
In the MPC-based approach, the controller seeks to optimize the control inputs (acceleration and steering angle) at each time step to minimize a given cost function. The cost function considers both the deviation from the reference trajectory and the proximity to obstacles, applying penalties when the vehicle deviates too far from the path or approaches an obstacle.


\subsubsection{Control Optimization}
The optimization process involves searching through a discrete set of possible control inputs for both acceleration $a_{\text{ref}}$ and steering angle $\delta_{\text{ref}}$, and finding the combination that minimizes the overall cost. This optimization process can be expressed as:
\begin{eqnarray} 
\min_{a_{\text{ref}}, \delta_{\text{ref}}} \sum_{i=1}^{N} \left[ \| \mathbf{s}_i - \mathbf{s}_i^{\text{ref}} \|^2 + \text{C}_{\text{obs}} + \text{C}_{\text{dist}} \right] \\
\text{C}_{\text{obs}} = w_{\text{obs}} \cdot \frac{1}{d_{\text{obs},i} + \epsilon} \cdot \theta_{\text{obs},i} \\
\text{C}_{\text{dist}} = w_{\text{dev}} \cdot \max(0, \| \mathbf{s}_i - \mathbf{s}_i^{\text{ref}} \| - d_{\text{max}})^2 
\end{eqnarray}

where:

- $\mathbf{s}_i$ and $\mathbf{s}_i^{\text{ref}}$ are the predicted and reference state at step $i$,

- $d_{\text{obs},i}$ is the distance to the nearest obstacle at step $i$,

- $\theta_{\text{obs},i}$ is the angle between the obstacle and the vehicle’s heading,

- $a_{\text{ref}}$ and $\delta_{\text{ref}}$ are the control inputs for acceleration and steering angle, respectively,

- $w_{\text{obs}}$ is the obstacle avoidance weight,

- $w_{\text{dev}}$ is the deviation penalty weight,

- $d_{\text{max}}$ is the maximum allowable deviation from the reference trajectory.

Optimization is performed by varying the control inputs acceleration $a_{\text{ref}}$ and steering angle $\delta_{\text{ref}}$ within their respective ranges, where acceleration 
$a_{\text{ref}} \in [-1, 1]$ and steering angle $\delta_{\text{ref}} \in \left[-\frac{\pi}{4}, \frac{\pi}{4}\right]$. For each combination of $a_{\text{ref}}$ and $\delta_{\text{ref}}$, the controller uses a kinematic model to predict the future trajectory over a prediction horizon of $N$ time steps. 

The predicted trajectory is evaluated using the previously defined cost function, which includes the Euclidean distance between the predicted state $\mathbf{s}_i$ and the reference state $\mathbf{s}_i^{\text{ref}}$, a penalty for proximity to obstacles that increases as the vehicle approaches obstacles within a safe distance, and a deviation penalty imposed if the predicted trajectory strays beyond the reference trajectory by a defined threshold $d_{\text{max}}$.

\subsubsection{Collision-Free Validation}
\label{subsec:Collision-Free Validation}

In each iteration, the predicted trajectory is validated to ensure that it does not lead the vehicle into a collision with obstacles. This is done by checking each predicted state in the trajectory to ensure it is collision-free. 


\subsubsection{Selecting the Best Control}

The goal of the optimization is to find the pair of control inputs $\left(a_{\text{ref}}, \delta_{\text{ref}}\right)$ that minimizes the total cost function. The control pair that yields the lowest cost is selected as the best control for the current time step:

\begin{eqnarray}
\left(a_{\text{best}}, \delta_{\text{best}}\right) = \arg\min_{a_{\text{ref}}, \delta_{\text{ref}}} J(a_{\text{ref}}, \delta_{\text{ref}})
\end{eqnarray}

Where $J(a_{\text{ref}}, \delta_{\text{ref}})$ represents the total cost for the predicted trajectory given the control inputs $a_{\text{ref}}$ and $\delta_{\text{ref}}$.

Bringing together both the optimization and the cost function:
\begin{eqnarray} 
\left(a_{\text{best}}, \delta_{\text{best}}\right) = \arg\min_{a_{\text{ref}}, \delta_{\text{ref}}} \sum_{i=1}^{N} \left[ \| \mathbf{s}_i - \mathbf{s}_i^{\text{ref}} \|^2 + \text{C}_{\text{obs}} + \text{C}_{\text{dist}} \right]
\end{eqnarray}

\subsection{Pure Pursuit}
A pure pursuit algorithm is enhanced with obstacle avoidance capabilities. The algorithm selects a target point on the reference trajectory, computes potential alternative paths to avoid obstacles, and selects the best obstacle-free path towards the goal.


\subsubsection{Target Selection}

\begin{align} 
    \mathbf{t} &= \arg\min_{i} \left( \text{distance}(\mathbf{s}, \mathbf{p}_i) \right) \\
    \nonumber\text{subject to} &~~~ \text{distance}(\mathbf{s}, \mathbf{p}_i) \geq \text{lookahead distance}
\end{align}

\subsubsection{Obstacle Avoidance}

Once the target state $\mathbf{t}$ is selected, the algorithm generates alternative target states to avoid potential obstacles. This is done by computing unit vectors perpendicular to the direction between the current vehicle position and the target, and adjusting the target position by a fixed distance in both directions. 

Let the direction vector from the current state to the target be $\mathbf{d} = (x_t - x, y_t - y)$. Two perpendicular unit vectors are computed as:
\begin{eqnarray} 
\mathbf{v}_1 = \left( -\frac{y_t - y}{\| \mathbf{d} \|}, \frac{x_t - x}{\| \mathbf{d} \|} \right), \quad \mathbf{v}_2 = \left( \frac{y_t - y}{\| \mathbf{d} \|}, -\frac{x_t - x}{\| \mathbf{d} \|} \right) 
\end{eqnarray}

For different adjustment distances $d_1, d_2, \dots, d_n$, the adjusted target states $\mathbf{t}_i$ are generated as:
\begin{eqnarray} 
\mathbf{t}_i = \mathbf{t} + d_i \cdot \mathbf{v}_1, \quad \mathbf{t}_{i+n} = \mathbf{t} + d_i \cdot \mathbf{v}_2 
\end{eqnarray}

The new candidate states $\mathbf{s}_i$ are then defined as:
\begin{align}
\mathbf{s}_i &= (x_i, y_i, \theta, v), \\
\nonumber\text{where} &~~~\text{v is velocity based on curvature}
\end{align}

The algorithm evaluates these candidate states to determine whether they are collision-free.

\subsubsection{Path Selection}

For each candidate adjusted state $\mathbf{s}_i$, the algorithm checks if it is collision-free. Among the collision-free states, the one that minimizes the distance to the goal is selected. This can be expressed as:
\begin{eqnarray} 
\mathbf{s}_{\text{best}} = \arg\min_{\mathbf{s}_i} \left( \text{distance}(\mathbf{s}_i, \text{goal}) \right)
\end{eqnarray}

If no collision-free state is found, the vehicle stays in place. 

This enhanced pure pursuit algorithm allows the vehicle to dynamically adjust its target to avoid obstacles while following a reference trajectory. The vehicle is able to safely navigate through a cluttered environment by selecting the best collision-free path at each step, ensuring it stays on track while avoiding obstacles.

\begin{figure*}[!h]
    \centering
    \subfigure[map\_easy]{
        \includegraphics[width=0.30\textwidth]{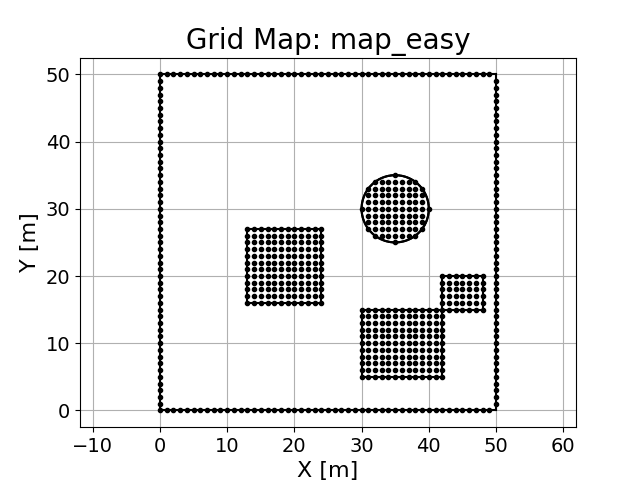}
        \label{fig:map_easy}
    }        
    \subfigure[map\_medium]{
        \includegraphics[width=0.30\textwidth]{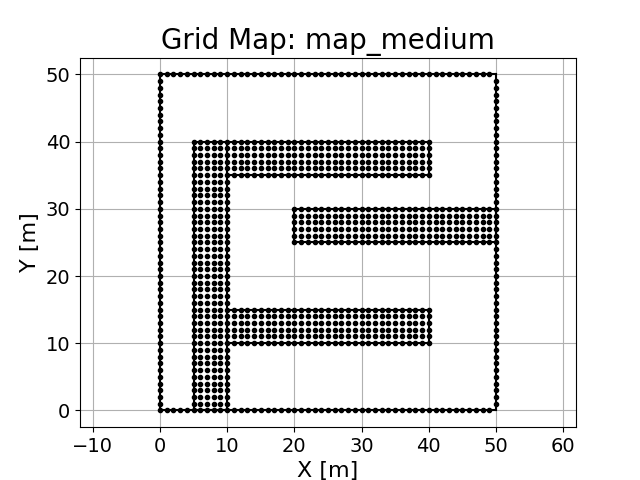}
        \label{fig:map_medium}
    }        
    \subfigure[map\_hard]{
        \includegraphics[width=0.30\textwidth]{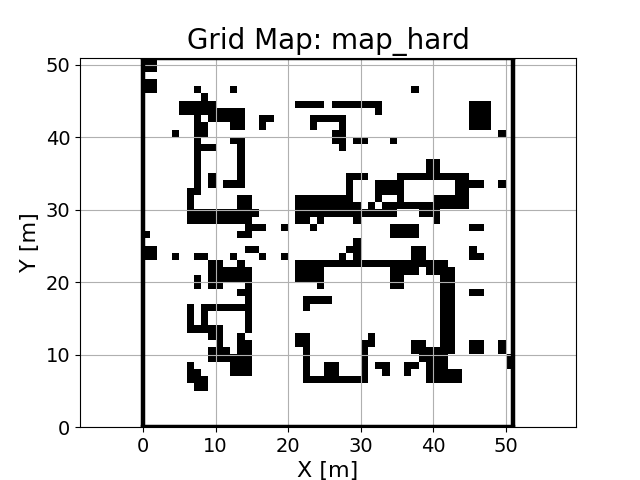}
        \label{fig:map_hard}
    }          
    \caption{Various Level of Map}
    \label{fig:map_base}
\end{figure*}

\subsection{InfoFusion Controller}

We introduce the InfoFusion Controller, which combines the strengths of MPC and Pure Pursuit control algorithms using MI. MI is used as the key factor in deciding how much to rely on each algorithm's state predictions in the final control decision.

\subsubsection{MI Calculation}

In our implementation, two sets of predicted states represent different states of the vehicle (e.g., position, heading angle, and speed) predicted by the MPC and Pure Pursuit controllers. These states are compared using histograms and their respective entropy values to compute the MI. This information is used to weigh the influence of each controller's predictions on the final decision.

\begin{algorithm}
\caption{MI-Based State Fusion}
\label{pseudoMI}
\begin{algorithmic}[1]
\State \textbf{Input:} ${\text{states}}\_{\text{pursuit}}$, ${\text{states}}\_{\text{mpc}}$, mi
\State \textbf{Output:} combined\_state
\State Initialize combined\_state as an empty array
\For{each state dimension $i$ in $\{x, y, \theta, velocity\}$}
    \If{$mi[i] > threshold$}
        \State Compute weight1 = $1 - \left( \frac{mi[i]}{mi[i] + 1} \right)$
        \State Compute weight2 = $\frac{mi[i]}{mi[i] + 1}$
        \State Set combined\_state[$i$] = weight1 $\cdot$ states\_pursuit[$i$] + weight2 $\cdot$ states\_mpc[$i$]
    \Else
        \State Set combined\_state[$i$] = states\_mpc[$i$]
    \EndIf
\EndFor
\State \Return combined\_state
\end{algorithmic}
\end{algorithm}

\subsubsection{Normalized MI}

Furthermore, we utilized the Normalized MI \cite{McDaid:2013, Estevez:2009} to enhance performance under varying conditions by ensuring that the influence of the two controllers is effectively balanced.

The Normalized MI is given by:
\begin{equation}
    I_{\text{norm}}(X, Y) = \frac{I(X; Y)}{\sqrt{H(X) H(Y)}}
\end{equation}

By using Normalized MI, the shared information between the controllers is scaled, allowing the influence of each controller to be better aligned and balanced, leading to more effective control decisions.

\subsubsection{MI-Based State Fusion}

Using MI, the predicted states from the two controllers are combined. When the MI between the controllers' predicted states is high, both controllers' predictions are weighted accordingly. If the MI is low, more reliance is placed on the MPC controller's predictions, as it generally provides more accurate and refined control.

The combined state $\mathbf{s}_{\text{combined}}$ is computed as:
\begin{eqnarray} 
    \mathbf{s}_{\text{combined}, i} = w_1 \cdot \mathbf{s}_{\text{pursuit}, i} + w_2 \cdot \mathbf{s}_{\text{mpc}, i}
\end{eqnarray}

where the weights $w_1$ and $w_2$ are derived based on the MI $I$:
\begin{eqnarray} 
    w_2 = \frac{I}{I + 1}, \quad w_1 = 1 - w_2
\end{eqnarray}

If the MI $I$ for a particular state dimension is low (below a threshold), the combined state relies more heavily on the MPC predictions:
\begin{eqnarray} 
    \mathbf{s}_{\text{combined}, i} = \mathbf{s}_{\text{mpc}, i} \quad \text{if} \quad I < 0.85
\end{eqnarray}

The InfoFusion Controller utilizes MI to intelligently combine the predictions of MPC and Pure Pursuit controllers. This ensures that the vehicle not only follows a desired trajectory but also effectively avoids obstacles by dynamically adjusting its reliance on each control algorithm based on their predictive performance.

\begin{figure} [!t]
\centering
\includegraphics[width=\linewidth]{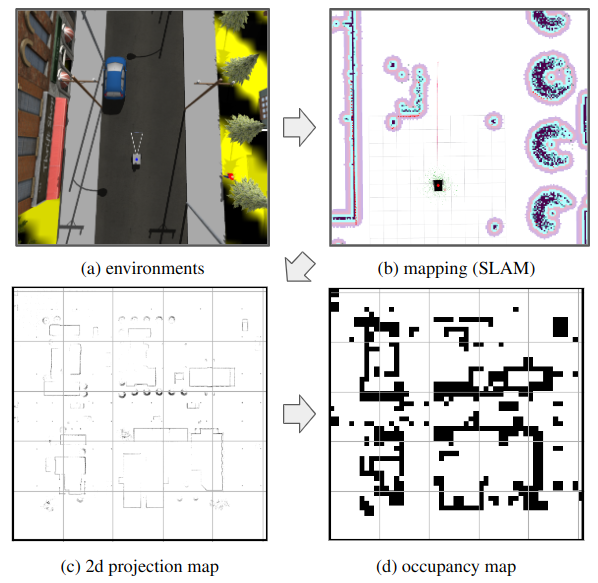}
\caption{How to generate occupancy map (map\_hard): Create a grid map by post-processing the map mapped to SLAM inside the RDSim simulation \cite{AuTURBO:2023}}\label{fig:generate_map_hard}
\end{figure}

\section{Experiment}
\label{sec:exp}

\subsection{Experimental Details}

All maps of Figure \ref{fig:map_base} were configured to a size of $50m \times 50m$, with each map designed to assess various terrain features. First, the map\_easy represents the simplest map, referencing the basic map used in the Informed TRRT Star paper. This map primarily focuses on evaluating the performance of fundamental path planning algorithms.

Next, map\_middle is a medium-difficulty map, designed to evaluate how well the algorithm can navigate through tight corners and narrow passages. This map is primarily suited for testing speed and handling stability.

Lastly, map\_hard is the most complex map, created using SLAM technology to simulate a real-world environment. While map\_easy and map\_medium are based on pre-designed obstacles and path structures, map\_hard, as shown in Figure \ref{fig:generate_map_hard}, was generated through SLAM to more realistically reflect actual driving environments. After generating the map, post-processing was done to make it suitable for path planning. This map allows for evaluating how efficiently and stably the path planning algorithm performs in complex environments.


\subsection{Overall Performance}

\begin{figure}[!t]
\centering
\includegraphics[width=\linewidth]{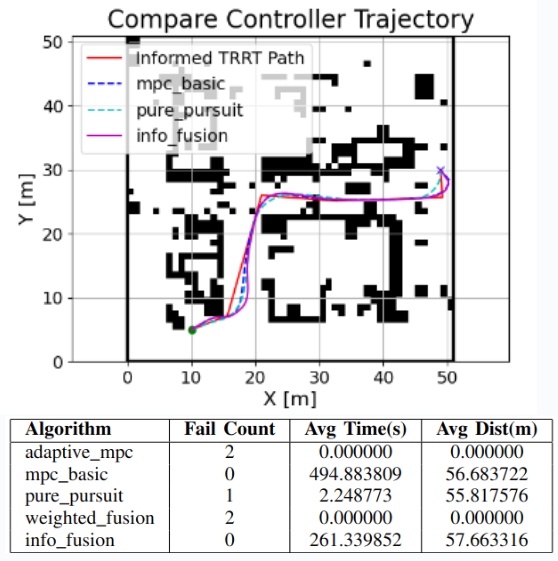}
\caption{Compare Overall Performance}
\label{fig:trajectory_hard}
\end{figure}

In the map\_hard environment in Figure \ref{fig:trajectory_hard}, where the map is complex and the paths are narrow, some algorithms encountered failures due to these challenging conditions. Among the algorithms that successfully navigated the environment, the proposed info\_fusion algorithm stands out. This algorithm not only maintains a safe distance from obstacles, similar to the mpc\_basic algorithm, but it also operates at twice the speed of mpc\_basic. This performance difference highlights the efficiency of info\_fusion, which offers faster and more effective path planning while preserving obstacle avoidance capabilities in complex environments.

\subsection{Safety Performance}

\begin{figure} [!h]
\centering
\includegraphics[width=\linewidth]{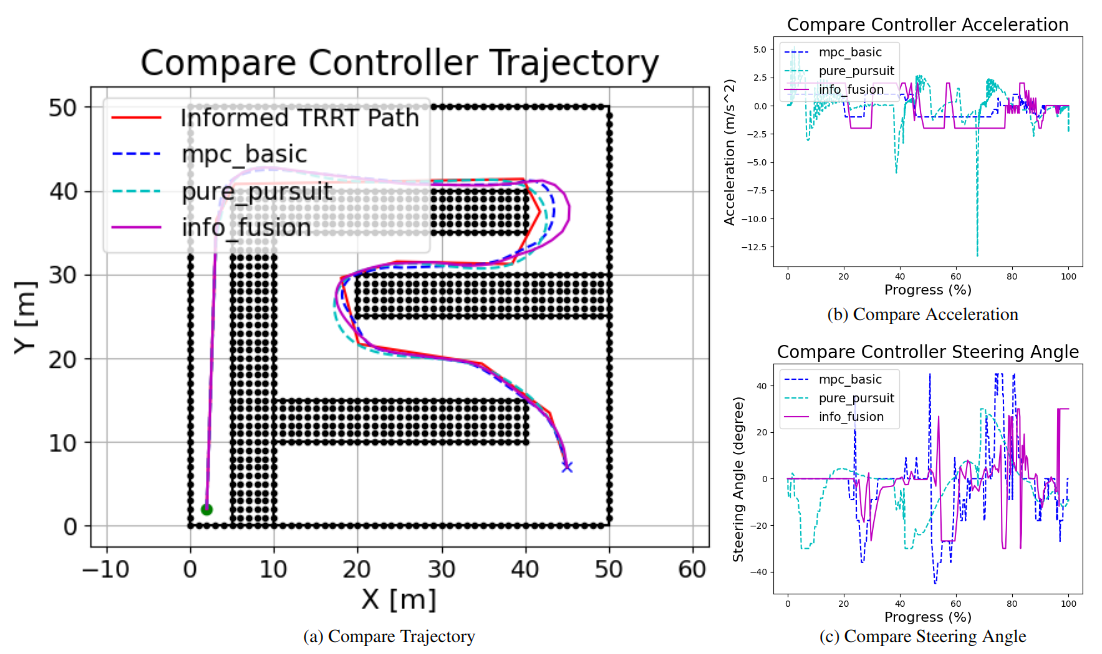}
\caption{Compare Safety Performance}\label{fig:fig_compare_safety}
\end{figure}

In order to compare the safety of the algorithms, we were conducted in map\_medium with a lot of cornering, and two key indicators were used. The first metric was acceleration, which indicates how stable the vehicle maintains its speed. The second metric was steering angle, which reflects the stability of the vehicle's handling.

Looking at acceleration in Figure \ref{fig:fig_compare_safety}, MPC attempts to keep control within the range of [-1 to 1], but as seen in the graph, it frequently oscillates back and forth excessively. This pattern can lead to unnecessary speed changes, reducing stability during driving. On the other hand, pure\_pursuit shows more stable acceleration without significant fluctuations, but it has a tendency to overly reduce speed before cornering, which leads to instability when transitioning from a straight to a curved path. In contrast, the info\_fusion algorithm demonstrates a more consistent pattern compared to MPC, with fewer sudden spikes, and it avoids the abrupt changes seen in pure\_pursuit, resulting in more stable speed control overall.

In terms of steering angle, MPC shows similar fluctuations as seen with acceleration, indicating reduced stability in handling. However, pure\_pursuit offers smoother and more stable handling during cornering. The info\_fusion algorithm combines the strengths of pure\_pursuit, maintaining relatively steady and stable steering angles with fewer abrupt turns. This results in improved handling stability compared to MPC, with a smoother driving experience that avoids unnecessary oversteering.

In conclusion, the info\_fusion algorithm demonstrates stable performance in both metrics, making it a safer and more reliable choice, particularly in scenarios where driving stability and safety are paramount.

\subsection{Obstacle Avoidance Performance}

\begin{figure} [!h]
\centering
\includegraphics[width=\linewidth]{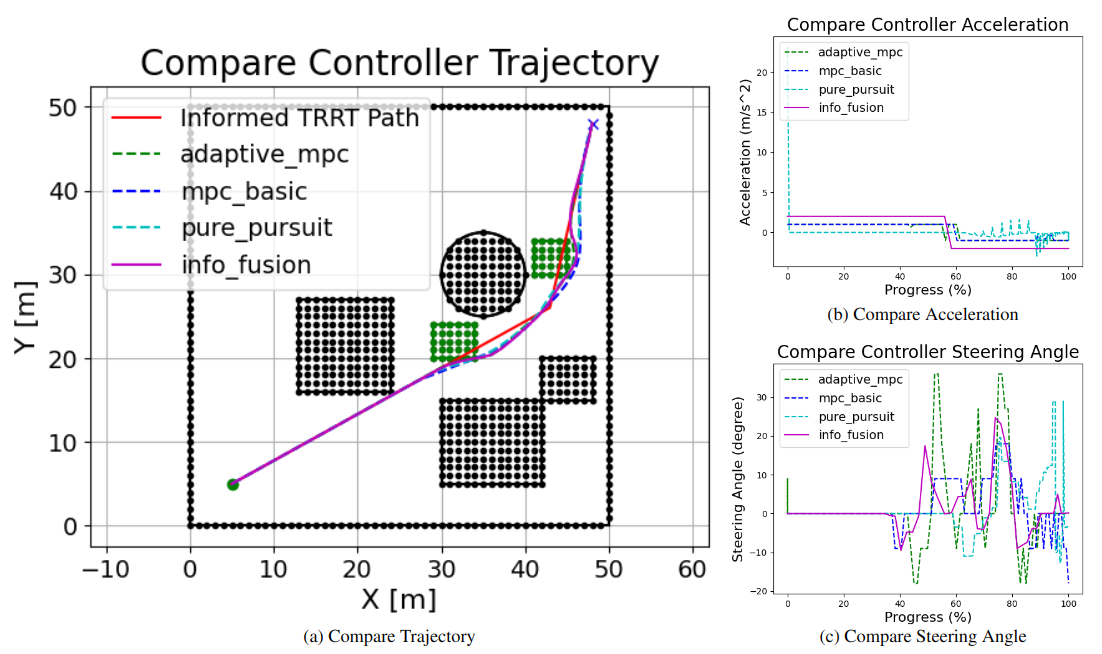}
\caption{Compare Obstacle Avoidance Performance}\label{fig:fig_compare_avoidance}
\end{figure}

In Figure \ref{fig:fig_compare_avoidance}, the InfoFusion algorithm successfully avoids dynamic obstacles while following the global path. The green boxes represent dynamic obstacles that appear during the route, requiring the controller to adapt accordingly.

In the left trajectory plot, InfoFusion demonstrates stable obstacle avoidance compared to other algorithms. While adaptive MPC and MPC basic show slight path distortions, and Pure Pursuit makes abrupt turns when navigating around dynamic objects, InfoFusion navigates smoothly with minimal deviation from the optimal path. This indicates that the InfoFusion controller maintains both safety and stability even when dynamic obstacles are introduced.

Turning to the Acceleration and Steering Angle plots on the right, InfoFusion exhibits less variability and more consistent performance than other algorithms. For instance, Pure Pursuit experiences sharp changes in acceleration when approaching obstacles, indicating sudden speed reductions. In contrast, InfoFusion maintains smoother and more consistent acceleration, even in the presence of obstacles. Additionally, its steering angle remains relatively stable compared to the more erratic behaviors of MPC basic and adaptive MPC, highlighting its superior handling stability.

These observations demonstrate that the InfoFusion algorithm reliably avoids dynamic obstacles while maintaining smooth speed control and stable handling, making it an effective and safe solution for real-world driving scenarios.


\section{Conclusion}
\label{sec:con}
In this paper, we introduced the InfoFusion Controller, a novel approach to autonomous path planning that integrates MPC and Pure Pursuit controllers using Mutual Information (MI). Leveraging the strengths of the Informed-TRRT* algorithm for global path planning and enhancing local performance through MI-based fusion, the controller demonstrates improved capabilities in complex and dynamic environments.

Our experimental results on various map complexities demonstrate that the InfoFusion Controller significantly improves safety, stability, and obstacle avoidance capabilities compared to traditional algorithms like MPC and Pure Pursuit alone. The controller consistently maintains stable behaviors and effectively navigates dynamic obstacles.



However, this study has limitations. The calculation of MI requires substantial computational resources, potentially impacting real-time performance in highly dynamic environments. Additionally, the controller's effectiveness depends on sensor data accuracy and model reliability; high sensor noise or model uncertainties may degrade performance.

Future work will focus on exploring more efficient algorithms or approximations to reduce the computational burden of MI calculations. Additionally, we plan to extend the InfoFusion Controller to more diverse and unpredictable real-world scenarios, integrating additional sensor inputs such as LIDAR and RADAR to improve robustness. Investigating adaptive weighting schemes within the MI framework will further enhance autonomous driving capabilities, and we will consider the effects of uncertainties in connected vehicle environments.




\begin{thebibliography}{00} 

\bibitem{SAE:J3016} SAE International, ``Taxonomy and Definitions for Terms Related to Driving Automation Systems for On-Road Motor Vehicles,'' \textit{SAE Standard J3016\_202104}, April 2021.

\bibitem{Li:2024} Y. Li, G. Li, and K. Peng, ``Research on Obstacle Avoidance Trajectory Planning for Autonomous Vehicles on Structured Roads,'' \textit{World Electr. Veh. J.}, vol. 15, no. 168, pp. 1–23, 2024.

\bibitem{Matsui:2022} N. Matsui et al., ``Local and Global Path Planning for Autonomous Mobile Robots Using Hierarchized Maps,'' \textit{J. Robotics and Mechatronics}, vol. 34, no. 1, 2022.

\bibitem{MacDonald:2016} R. A. MacDonald and S. L. Smith, ``Reactive Motion Planning in Uncertain Environments via Mutual Information Policies,'' \textit{WAFR}, 2016.

\bibitem{Yang:2022} W. Yang, C. Li, and Y. Zhou, ``A Path Planning Method for Autonomous Vehicles Based on Risk Assessment,'' \textit{World Electr. Veh. J.}, vol. 13, pp. 234, 2022.

\bibitem{Daniel:2010} K. Daniel, A. Nash, S. Koenig, and A. Felner, ``Theta: Any-Angle Path Planning on Grids,'' \textit{Journal of Artificial Intelligence Research}, vol. 39, pp. 533-579, 2010.

\bibitem{Shi:2024} C. Shi and Z. Wu, ``Informed-TRRT*: An Improved Sampling-Based Path Planning Algorithm,'' \textit{Discrete and Continuous Dynamical Systems - Series S}, 2024.

\bibitem{Lee:2024} T. Lee and Y. Jeong, ``A Tube-Based Model Predictive Control for Path Tracking of Autonomous Articulated Vehicle,'' \textit{Actuators}, vol. 13, no. 5, pp. 164, 2024.

\bibitem{Li:2023} B. Li et al., ``Adaptive Pure Pursuit: A Real-time Path Planner Using Tracking Controllers,'' \textit{IEEE Transactions on Intelligent Vehicles}, vol. 13, 2023.

\bibitem{Singh:2024} R. Singh, ``A Hybrid Path Planning Technique for the Time-Efficient Navigation of Unmanned Vehicles in an Unconstrained Environment,'' \textit{Journal of Intelligent \& Robotic Systems}, vol. 110, pp. 119, 2024.

\bibitem{Hadizadeh:2024} H. Hadizadeh, S. F. Yeganli, B. Rashidi, and I. V. Bajic, ``Mutual Information Analysis in Multimodal Learning Systems,'' \textit{IEEE International Conference on Multimedia Information Processing and Retrieval (MIPR)}, 2024.

\bibitem{He:2020} X. He and X. Ji, ``Multi-Modal Vehicle Trajectory Prediction Based on Mutual Information,'' \textit{IET Intelligent Transport Systems}, vol. 14, no. 2, pp. 105-114, 2020.

\bibitem{Kopylova:2008} Y. Kopylova, D. A. Buell, C. T. Huang, and J. Janies, ``Mutual Information Applied to Anomaly Detection,'' \textit{Journal of Communications and Networks}, vol. 10, no. 1, pp. 52-59, 2008.

\bibitem{Bai:2021} S. Bai, T. Shan, F. Chen, L. Liu, and B. Englot, ``Information-Driven Path Planning,'' \textit{Current Robotics Reports}, vol. 2, no. 3, pp. 177–188, 2021.

\bibitem{Yin:2023} J. Yin and W. Fu, ``A Safety Navigation Method Integrating Global Path Planning and Local Obstacle Avoidance for Self-Driving Cars in a Dynamic Environment,'' \textit{Science China Information Sciences}, vol. 28, no. 3, pp. 1318–1328, 2023.

\bibitem{McDaid:2013} A. F. McDaid, D. Greene, and N. Hurley, ``Normalized mutual information to evaluate overlapping community finding algorithms,'' \textit{arXiv}, vol. 1110.2515v2, 2013. Available: https://github.com/aaronmcdaid/Overlapping-NMI

\bibitem{Estevez:2009} P. A. Estévez, M. Tesmer, C. A. Perez, and J. M. Zurada, ``Normalized mutual information feature selection,'' \textit{IEEE Transactions on Neural Networks}, vol. 20, no. 2, pp. 189–201, 2009.

\bibitem{AuTURBO:2023} AuTURBO, ``RDSim: Autonomous Vehicle Simulation for Obstacle Avoidance,'' \textit{GitHub repository}, Available: \url{https://github.com/AuTURBO/RDSim}, 2023.








\end{thebibliography}
\end{document}